\documentclass[sigconf,authorversion,nonacm]{acmart}
\AtBeginDocument{%
  }
\usepackage{tabularx}
\setcopyright{acmlicensed}
\copyrightyear{2018}
\acmYear{2018}
\acmDOI{XXXXXXX.XXXXXXX}

\acmConference[Conference acronym 'XX]{Make sure to enter the correct
  conference title from your rights confirmation email}{June 03--05,
  2018}{Woodstock, NY}

\acmISBN{978-1-4503-XXXX-X/2018/06}

\usepackage{xspace}
\usepackage[symbol]{footmisc}
\usepackage{amsmath}

\newcommand{\ft}{\textsc{$\text{FastTrack}$}\xspace}
\newcommand{\fa}{\textsc{FullAgentic}\xspace}
\newcommand{\point}[1]{\vspace{.05in} \par\noindent\textbf{#1}. }

\begin{document}

\title{Compliance Brain Assistant: Conversational Agentic AI for Assisting Compliance Tasks in Enterprise Environments}


\author{Shitong Zhu*, Chenhao Fang*, Derek Larson*, Neel Reddy Pochareddy*, Rajeev Rao*, Sophie Zeng*, Yanqing Peng*, Wendy Summer*, Alex Goncalves, Arya Pudota, Hervé Robert}
\affiliation{%
  \institution{Meta}
  \city{Menlo Park}
  \state{California}
  \country{USA}
}

\renewcommand{\shortauthors}{Zhu et al.}

\begin{abstract}
This paper presents Compliance Brain Assistant (CBA), a conversational, agentic AI assistant designed to boost the efficiency of daily compliance tasks for personnel in enterprise environments. To strike a good balance between response quality and latency, we design a \textit{user query router} that can intelligently choose between (i) \ft: to handle simple requests that only need additional relevant context retrieved from knowledge corpora; and (ii) \fa: to handle complicated requests that need composite actions and tool invocations to proactively discover context across various compliance artifacts, and/or involving other APIs/models for accommodating requests -- a typical example would be to start with a user query, use its description to find a specific entity and then use the entity's information to query other APIs for curating and enriching the final AI response.

Our experimental evaluations compared CBA against an out-of-the-box LLM on various real-world compliance-related queries targeting various personas. We found that CBA substantially improved upon the vanilla LLM's performance on metrics such as average keyword match rate (83.7\% vs. 41.7\%) and LLM-judge pass rate (82.0\% vs. 20.0\%), highlighting the effectiveness of our compliance-oriented enhancements. We also compared metrics for the full routing-based design against the \ft-only and \fa modes and found that it had a better average match-rate and pass-rate while keeping the response latency approximately the same. This finding validated our hypothesis that the routing mechanism leads to a good trade-off between the two worlds.
\end{abstract}




\maketitle

\footnotetext[1]{Authors contributed equally}

\section{Introduction}

In today's world, organizations are required to adhere to a diverse array of requirements, best practices, and industry-specific regulations from multiple sources. To effectively manage compliance tasks, organizations assign dedicated personnel with expertise in compliance to oversee these responsibilities. Throughout this process, these experts often benefit from the assistance of conversational chatbots. By providing precise answers and clarifications on complex subjects, chatbots can significantly enhance both the efficiency and quality of task completion.

In the past, pattern-matching schemes and conventional machine-learning classifiers have been used to categorize such questions into predefined labels and provide standard responses or documents. However, these questions often come in the form of free-flowing prose, blending complex vocabulary, engineering jargon, and other terminology common in internal discussions. The complexity and diversity of these questions make it challenging to be answered by matching to preexisting responses only -- in many instances, the provided answer may not relate to the question, or it may be buried within a lengthy document that the user must sift through. Worse, if the question contains ambiguities, the response may not address the engineer's actual concerns, which they might not even realize due to lacking background knowledge in the space.

\begin{figure}
    \centering
    \includegraphics[width=\linewidth]{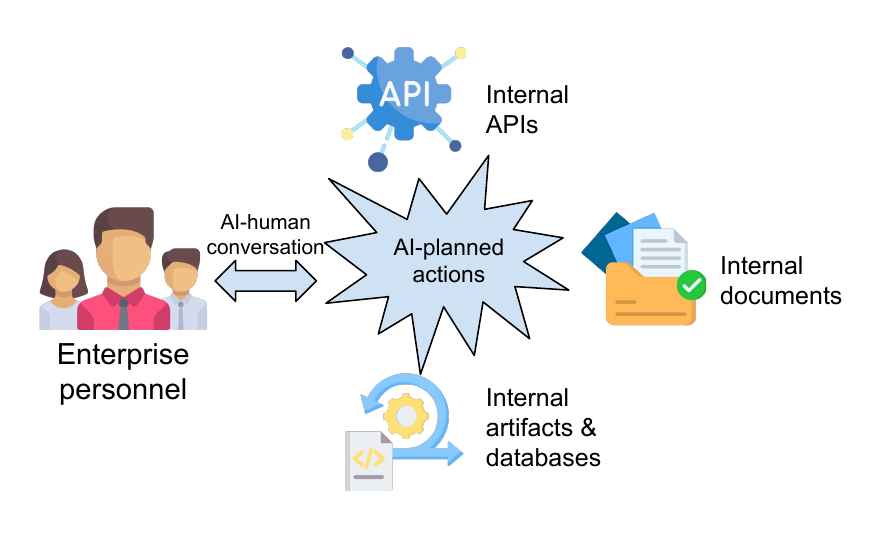}
    \caption{Interaction paradigm for enterprise personnel to be assisted by conversational AI}
    \label{fig:ideal_paradigm}
\end{figure}

We are looking for a smarter AI system that can assist these enterprise personnel better in completing the compliance tasks. Ideally, it should understand the semantics of a question, clarify ambiguous phrasing, provide precise answers, retain context across interactions, and tailor explanations to the engineer's expertise level. More importantly, the assistant should support task owners by invoking various enterprise-internal APIs to incrementally gather necessary information during the process of answering user queries. We depict this ideal interaction paradigm in Figure~\ref{fig:ideal_paradigm}.
%

The recent advancements in Large-Language Models (LLMs) research, which combine extensive world knowledge with the ability to follow instructions and perform ``actions'' that interact with external APIs, have made it technically feasible to develop such a conversational, agentic assistant. By leveraging LLM as the core reasoning engine and equipping it with knowledge base, memory and tools, we can address this long-standing challenge in the compliance space for enterprise bodies. 
%

Developing such a system is not straightforward. Vanilla LLM models such as GPT~\cite{grattafiori2024llama}, are trained on a broad Internet corpus and therefore lack the knowledge regarding enterprise-internal compliance-related documents, policies specific to certain domains, as well as real-time information from artifacts (e.g., reviews) and APIs from enterprise databases, that a compliance task often depends on. While fine-tuning models for specific tasks has shown promising results~\cite{zhang2023instruction}, these models often risk losing their general reasoning capabilities, known as the catastrophic forgetting problem~\cite{luo2023empirical}, and also cannot access real-time information from enterprise databases. Furthermore, while integrating vanilla LLMs with Retrieval-Augmented Generation (RAG) systems for information retrieval can be helpful for them to retrieve context in a real-time manner and answer simple domain-specific queries~\cite{fang2024ingest}, it is oftentimes insuffient for end-to-end enterprise compliance tasks, because these tasks could require multiple steps of actions, to sequentially gather required information (which a single-step RAG retrieval is not enough).

To address these challenges, we developed Compliance Brain Assistant (CBA), a cascading system leveraging LLMs, RAG and agentic tools and intelligently orchestrate them for best responding to user queries. Specifically, a lightweight classifier (the router) inspects the incoming user query and determines the most efficient workflow that yields high-quality answers. The router itself relies on the LLM to pick the most suitable workflow, by considering factors such as the question's complexity, the availability of relevant tools and models, as well as a predefined suite of example user queries suiting different workflows.
Concretely, there are two paths of workflows that the router selects between:
%
\begin{itemize}
    \item \textbf{\ft: LLM response with optional context retrieval.} When the router determines that the question is broadly generic (i.e., no RAG is needed and one call to out-of-box LLM is enough), or only needs one-step context retrieval via RAG (depending on the user query contents), it routes the query to this workflow. This path involves no external tool calls, so it has relatively low latency.
    \item \textbf{\fa: multi-step autonomous context gathering with specialized tools.} If the query is determined to require chain of actions to acquire additional context from compliance artifacts in enterprise databases, the router sends the query to the agent engine, which can call any tool exposed by the tool catalogue, including but not limited to:
        \begin{itemize}
            \item Searching for related artifacts using a RAG system
            \item Fetching metadata or details of these related artifacts
            \item Invoking fine-tuned specialist models trained exclusively on high-traffic sub-domains (e.g., data retention policies, cross-border data transfers), which demonstrate significantly better performance than generalist models on these specific tasks
        \end{itemize}
    The agent iteratively gathers snippets, verifies the content, and assembles an evidence set that is passed back to the LLM to draft a grounded answer. By using specialist models trained on specific sub-domains, we can ensure that the CBA provides highly accurate and relevant answers, even for complex and nuanced questions. These models have been shown to outperform general-purpose models on tasks such as data retention policy analysis and cross-border data transfer guidance.
\end{itemize}

\section{Architecture}
\begin{figure}
    \centering
    \includegraphics[width=\linewidth]{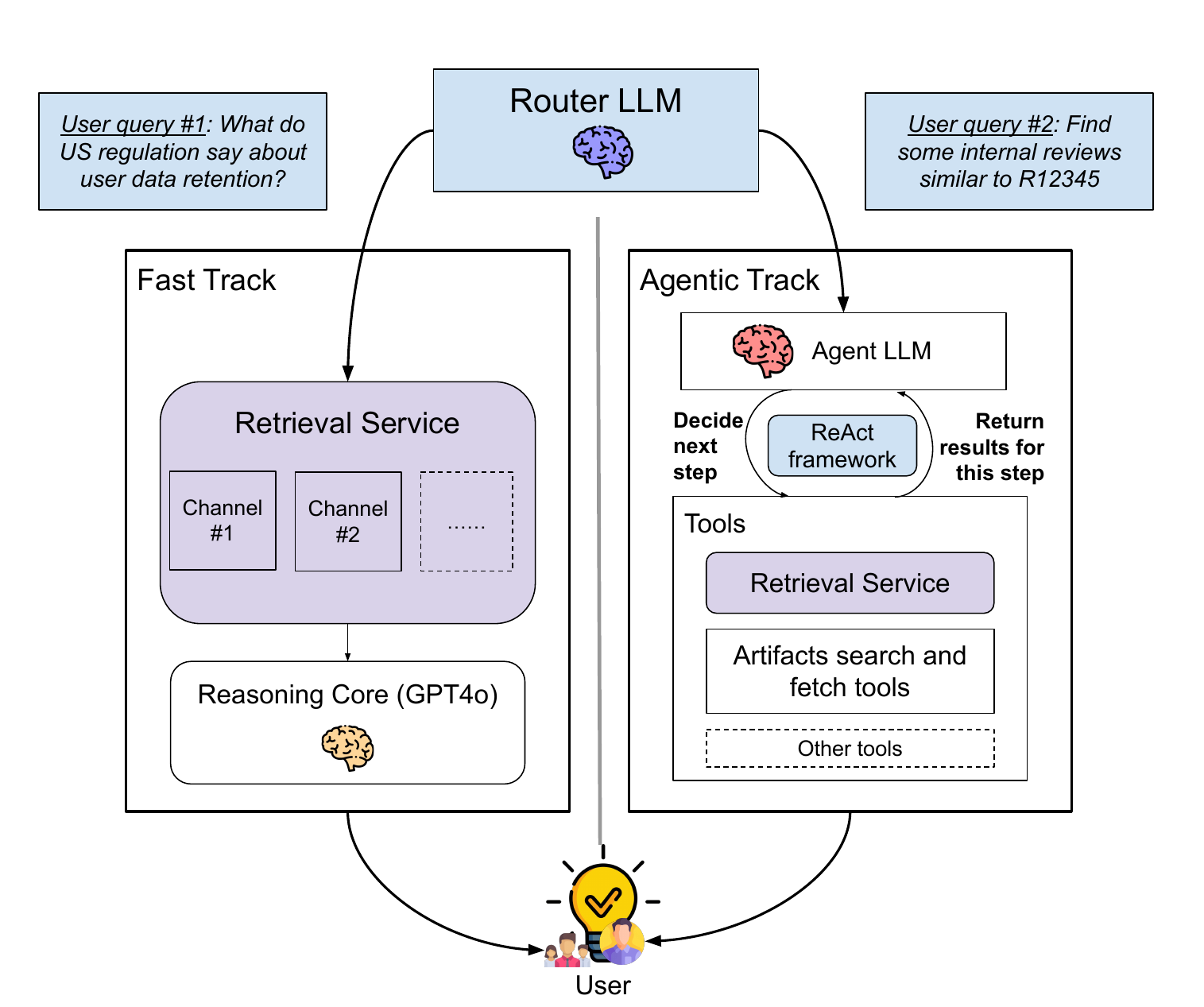}
    \caption{CBA architecture with main building blocks}
    \label{fig:arch}
\end{figure}
The architecture of CBA is designed to address the speed and quality challenges described in the section above. It consists of three key components: the Router, \ft flow, and \fa flow. Each component plays a important role in optimizing the system's performance and user experience.

\vspace{2mm}
\textbf{Router.} At the core of the new architecture is a routing-based framework that intelligently directs user queries to the appropriate processing track. The Router leverages an efficient while effective LLM model, Llama3.1-8B, to determine whether a query requires the full agentic workflow or can be more efficiently handled by the \ft flow. This decision-making process is important for reducing latency by avoiding unnecessary steps for queries that do not require complex reasoning or multi-step tool use. 

To enable the model to function effectively as a query classifier, we carefully crafted a prompt that describes the circumstances under which \ft should be employed, and vice versa. The criteria are straightforward: if a query requires information about any internal artifacts, it should be directed to the agentic track; otherwise, \ft is sufficient. This description alone can accurately route most queries, but to further enhance accuracy, we sampled a selection of short queries from real-world traffic. These queries were processed through the agentic flow, and we labeled them based on whether any tool was invoked. Currently, we have 10 queries included in the prompt as in-context learning, but we may add or update the examples as we receive more feedback from our production traffic. As we will demonstrate in our experiments, our current model achieves high accuracy while maintaining low latency.

\vspace{2mm}
\textbf{\ft flow.} \ft is tailored for conversations that request AI to answer questions based on information from compliance knowledge bases, specifically those unrelated to detailed instances of enterprise-internal artifacts (e.g., understanding what entity-X pertains to). By employing a fast Retrieval-Augmented Generation (RAG)~\cite{gao2023retrieval} combined with an LLM, this flow significantly reduces response times, providing users with swift and relevant answers. 
The RAG module, referred to as Retrieval Services, encompasses multiple retrieval sources, including enterprise-internal search results from a vast array of non-compliance-specific wiki pages or posts, as well as proprietary documents with advanced data access controls. These indexes are prepared using semantic chunking algorithms and quality safeguards to filter out low-quality pages and chunks, ensuring that only the most relevant and high-quality information is retrieved. 
When entering the \ft flow, RAG was first executed to pull information from a variety of sources described above. The retrieved information was integrated to the prompt to allow the LLM to augment its understanding of the query context and to generate a more accurate and informative response.
By streamlining the retrieval process and leveraging advanced indexing techniques, the \ft flow enhances the system's ability to deliver timely and accurate responses, thereby improving the overall user experience.

\vspace{2mm}
\textbf{\fa flow.} For queries requiring tool use and comprehensive reasoning over existing enterprise database entities, the \fa flow is employed. This flow utilizes a catalog of tools in these main types: (1) artifact fetching tools that access contents/details of enterprise-internal artifact using entity locators (e.g., IDs); (2) artifact semantic search tools that find entities related to the user query, or similar entities to the ones mentioned in the conversation; (3) knowledge retrieval tools calling the Retrieval Service also used in the \ft flow, to enable combined context from knowledge and artifacts; and (4) specialized AI models for specific tasks (e.g., fine tuned LLMs). To manage the growing number of tools and ensure accuracy and efficiency in the agent's tool use planning, the tools were implemented with consistent interfaces, and organized properly to minimize tool planning redundancy. In addition, concurrent tool uses were enabled in certain cases (when two tool operations do not depend on each other) to reduce the overall latency. 

\begin{figure}
    \centering
    \includegraphics[width=\linewidth]{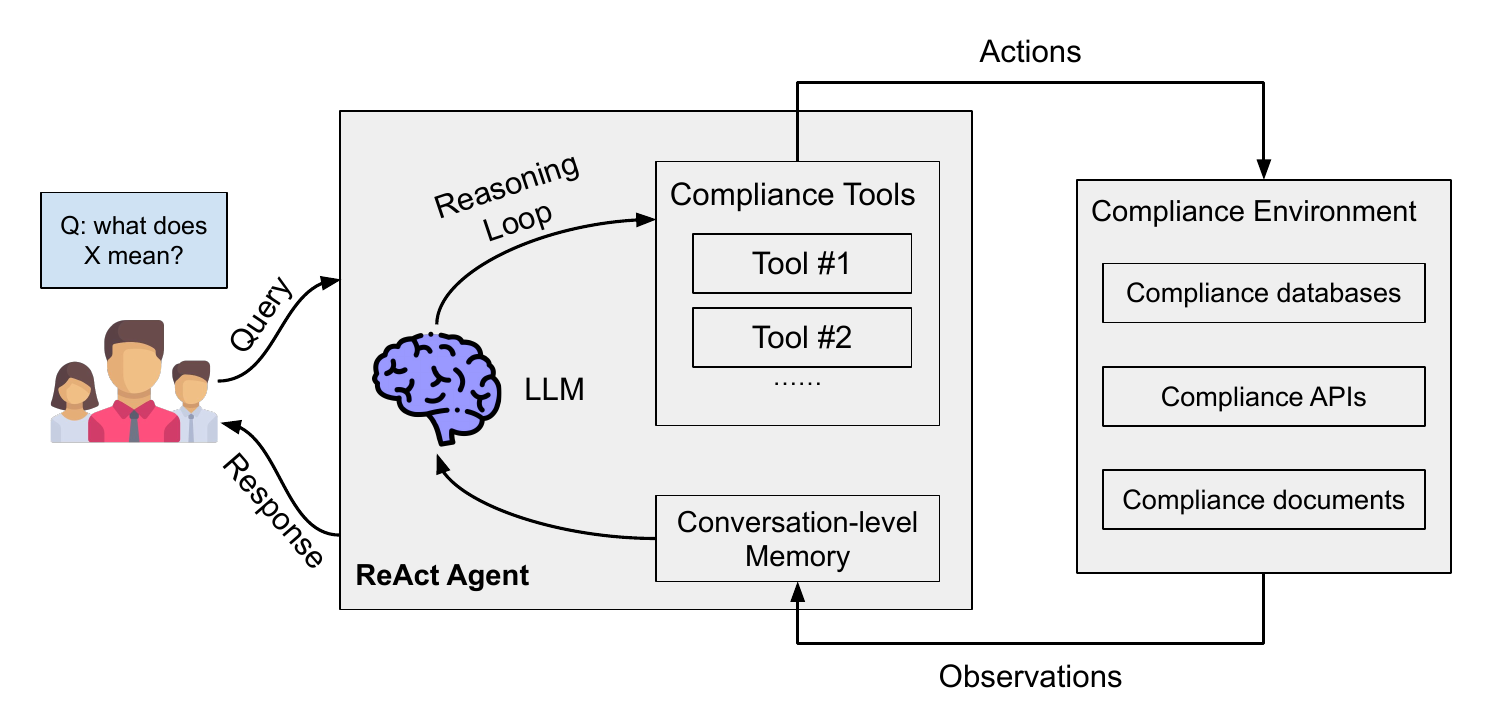}
    \caption{Customized ReAct-style compliance agent framework used in \fa}
    \label{fig:react_agent}
\end{figure}

In the \fa flow, LLM acts as a ReAct agent~\cite{yao2023react} that orchestrates the use of various tools to address complex queries requiring comprehensive reasoning over existing enterprise entities. In a nutshell, ReAct (Reasoning and Acting) paradigm (as depicted in Figure~\ref{fig:react_agent}) is an advanced framework for large language model (LLM) agents that integrates both chain-of-thought reasoning and action execution in an interleaved fashion. Unlike traditional LLM prompting techniques that separate reasoning from interaction with external tools, ReAct enables agents to alternately generate intermediate reasoning steps and invoke actions (e.g., search queries, environment manipulations, or API calls) as part of a single coherent response trajectory. This joint formulation allows the agent to dynamically update its reasoning based on observed outcomes of prior actions, facilitating more robust problem solving and decision making in interactive or information-seeking tasks. Empirical results show that ReAct-style agents outperform purely reactive or purely deliberative baselines in tasks requiring multi-step inference and external information retrieval, demonstrating the advantages of tightly coupling reasoning and action within LLM-based agents.
For \fa, we further adapted ReAct to fit unique characteristics of compliance tasks (e.g., usually these tasks require a mix of internal APIs, databases and documents to operate among, unlike general tasks that are more homogeneous). Specifically, CBA agent is designed with the following steps.
\begin{enumerate}
    \item \textbf{Compliance tool initialization}: Upon entering the agentic flow, the LLM is provided with a list of available tools, each accompanied by a detailed description and example usage scenarios. This setup allows the LLM to understand the capabilities and limitations of each tool, enabling it to make informed decisions about which tools to employ for a given query.
    \item \textbf{Compliance reasoning and planning}: The LLM begins by analyzing the user query to determine the necessary steps to generate an accurate response. It reasons about the query's requirements and decides which compliance tool or combination of tools will best address the query. This decision-making process involves evaluating the nature of the query and the type of information or action required. We tailor the guiding prompt used in this step to define criteria and instructions that helps LLM make correct tool use plans for common compliance tasks, mimicking thinking processes of expert compliance personnel.
    \item \textbf{Compliance tool execution}: In each interaction turn, the LLM selects and executes one tool from its toolkit. The LLM then processes the output from the tool to extract relevant information.
    \item \textbf{Iterative reasoning}: After obtaining the tool's output, the LLM reasons on the results to determine the next course of action. It assesses whether additional information is needed or if the current data suffices to formulate a response. Based on this assessment, the LLM may choose to execute another tool, refine its search, or proceed to generate a response for the user.
    \item \textbf{Response generation}: Once the LLM has gathered sufficient information and completed its reasoning process, it compiles the findings into a coherent and insightful response for the user. This response is designed to be both accurate and actionable, addressing the user's query with precision and depth.
\end{enumerate}

This design allows \fa to offer an adaptive and efficient workflow that caters to the diverse needs of its user base while maintaining high-quality responses. By leveraging a comprehensive suite of tools and advanced reasoning capabilities, this flow ensures that even the most challenging queries are addressed with precision and depth, providing users with insightful and actionable responses.

\section{Evaluations}
\subsection{Router Accuracy}

To validate the router's ability to correctly direct queries to the appropriate workflow path (\ft or \fa), we conducted a controlled experiment using historical production data. This evaluation helps ensure the router's reliability as the decision-making core of the CBA system.

\subsubsection{Data Collection}
We were inspired by common user questions in compliance space, and summarized 15 representative queries as the evaluation samples. These queries were manually labeled by experts as requiring:
\begin{itemize}
    \item \ft: Generic questions answerable via RAG-based context enhancements only
    \item \fa: Questions requiring company-specific context or domain expertise
\end{itemize}

\subsubsection{Test Methodology}
The router was presented with the 15 queries in isolation (without prior conversation context) and forced to choose between \ft or \fa. We compared its decisions against the expert labels using standard classification metrics. We tested different sizes of Llama 3 series models (3.1-8B, 3.2-1B, 3.2-3B, 3.3-70B) to compare their efficiency and effectiveness.

\subsubsection{Results}
\begin{table}[h]
\centering
\renewcommand{\arraystretch}{1.5}

\caption{Router accuracy and efficiency results: (a) confusion matrix, (b) classification metrics and (c) latency}
\begin{minipage}{0.45\textwidth}
    \centering
    \begin{tabular}{|c|c|c|c|c|}
        \hline
        \textbf{Actual/Predicted} & \textbf{1B} & \textbf{3B}  & \textbf{8B} & \textbf{70B} \\
        \hline
        \textbf{\ft/\ft} & 4 & 7 & 7 & 7\\
        \hline
        \textbf{\ft/\fa} & 4 & 1 & 1 & 1\\
        \hline
        \textbf{\fa/\ft} & 2 & 1 & 1 & 1\\
        \hline
        \textbf{\fa/\fa} & 5 & 6 & 6 & 6\\
        \hline
    \end{tabular}
    \vspace{1mm}
    \caption*{(a) Confusion matrix}
\end{minipage}
\hfill
\begin{minipage}{0.45\textwidth}
    \centering
    \begin{tabular}{|c|c|c|c|c|}
        \hline
        \textbf{Class} & \textbf{1B} & \textbf{3B} & \textbf{8B} & \textbf{70B}\\
        \hline
        \textbf{\ft Precision} & 66.7\% & 87.5\%  & 87.5\% & 87.5\%\\
        \hline
        \textbf{\ft Recall} & 50.0\% & 87.5\%  & 87.5\% & 87.5\%\\
        \hline
        \textbf{\fa Precision} & 55.6\% & 85.7\% & 85.7\% & 85.7\% \\
        \hline
        \textbf{\fa Recall} & 71.4\% & 85.7\% & 85.7\% & 85.7\% \\
        \hline
    \end{tabular}
    \vspace{1mm}
    \caption*{(b) Classification metrics}
\end{minipage}
\hfill
\begin{minipage}{0.45\textwidth}
    \centering
    \begin{tabular}{|c|c|c|c|c|}
        \hline
        \textbf{Model} & \textbf{1B} & \textbf{3B} & \textbf{8B} & \textbf{70B}\\
        \hline
        \textbf{Latency (s)} & 0.77 & 0.88  & 1.61 & 3.15\\
        \hline
    \end{tabular}
    \vspace{1mm}
    \caption*{(c) Average Latency per Request}
\end{minipage}
\label{tbl:1}
\end{table}

The results are presented in Table~\ref{tbl:1}. Given the relatively straightforward nature of this classification task for LLMs, any Llama model above 3B parameters performs effectively. Among these, the router demonstrates strong performance (86.7\% overall accuracy) in distinguishing between generic and context-rich queries, validating its role as the decision-making backbone of CBA. The errors mostly arise from borderline cases (e.g., queries that can be considered to only marginally benefit from having artifact support), and reveal clear improvement opportunities - we have added these misclassified examples to our in-context dataset and will retrain the router periodically using this growing collection of edge cases. This iterative refinement process will enable continuous improvement in routing accuracy while maintaining low latency through the lightweight classifier architecture. We have selected the Llama3.1-8B model as the base for the router due to its strong performance, sufficient speed for our use case, and greater potential for future enhancements compared to the smaller 3B model.
Future work will explore incorporating real-time feedback from privacy specialists to further enhance decision quality.

\subsection{End-to-end Performance}

To understand how CBA performs with the router, we conducted a comprehensive experiment to evaluate the performance of different workflows on answering privacy-related tasks with regards to the accuracy and latency. The experiment consisted of four main conditions: 
\begin{enumerate}
    \item  \textbf{Vanilla LLM}: This condition used only Vanilla LLM, a general-purpose language model, without any additional agentic workflows or RAG enhancements.
    \item  \textbf{\ft}: In this condition, we utilized the \ft workflow of the CBA, which is optimized for speed and efficiency. 
    \item  \textbf{\fa}: Here, we employed CBA and always use the agentic flow.
    \item  \textbf{Routing-based}: This condition used CBA with the Router module based on Llama3.1-8B as previously described, which dynamically selects the most suitable workflow (\ft or \fa) based on the input prompt.
\end{enumerate}

\subsubsection{\textbf{Dataset}}
 To best understand different aspects of capabilities of CBA agent, we curated three datasets:
    \begin{enumerate}
        \item \textbf{Compliance Knowledge Benchmark}: includes 50 freeform QA samples to evaluate how well AI systems understand compliance concepts (e.g., questions related to data ownership).
        
        \item \textbf{Regulation Knowledge Benchmark}: includes 14 free-form Q\&A samples to assess questions related to Government regulations. This is a relatively difficult benchmark due to the ambiguous nature of the expected responses.
        
        \item \textbf{Compliance Artifact Understanding Benchmark}: includes 54 questions that requires the agentic tools from CBA to interact with compliance-related artifacts to locate and digest the correct information (for example "who is the owner for a given compliance artifact").
    \end{enumerate}
We constructed the Compliance and Regulation Knowledge Benchmark with common questions on compliance and regulation with reference freeform answers for each question as well as keywords for each questions that the answer should contain. For the Privacy Artifact Understanding Benchmark, we construct it with common queries on privacy artifacts and provided only the keyword answers for each question.

\subsubsection{\textbf{Evaluation Method and Metrics}}
To assess response quality from all setups, we use graders utilizing LLM-as-a-Judge~\cite{zheng2023judging} to grade questions that have free text ground truth answers, and use keyword-matching mechanism to assess on the provided keyword ground truth on the evaluation datasets, with global and average keywords. Latency is another aspect that we measure besides response accuracy.  
We therefore report the \emph{average per-sample latency} (in seconds) for each
dataset, i.e., the mean time it takes the system to answer one question.
The metrics we track with above measuring methods are as follows:

\point{Global Match Rate} The percentage of correct matches between the model's output and ground truth answers, calculated as the ratio of matched keywords from the answer with respect to all keywords for all questions in the dataset.

\begin{align}
\mathrm{GlobalMatchRate}
  &= \frac{\displaystyle
           \sum_{i=1}^{N}
           \sum_{k \in K_i} \mathbf{1}\!\bigl\{k \in M_i\bigr\}}
          {\displaystyle
           \sum_{i=1}^{N} |K_i|}
\end{align}
\noindent\textit{where}  
\(N\) is the total number of evaluation questions;  
\(K_i\) is the set of ground-truth keywords for question \(i\) (\(|K_i|\!:=\) its size);  
\(M_i\) is the model’s answer to question \(i\); and  
\(\mathbf{1}\{\cdot\}\) is the indicator function (1 if the condition is true, 0 otherwise).

\point{Average Match Rate} The average match rate across all test cases, computed by calculating the keyword match rate for each question and then taking the mean across all questions in the dataset.

\begin{align}
\mathrm{AverageMatchRate}
  &= \frac{1}{N}\;
     \sum_{i=1}^{N}
       \frac{\displaystyle
             \sum_{k \in K_i} \mathbf{1}\!\bigl\{k \in M_i\bigr\}}
            {|K_i|}
\end{align}
\noindent\textit{where}  
\(N\), \(K_i\), \(|K_i|\), \(M_i\), and \(\mathbf{1}\{\cdot\}\) are already defined as above.

\point{Pass Rate} The percentage of test cases where the model's output meets or exceeds the required accuracy threshold based on the grader utilizing LLM-as-a-Judge mechanism.

\begin{align}
\mathrm{PassRate}
  &= \frac{1}{N}\;
     \sum_{i=1}^{N}
       \mathbf{1}\!\bigl\{\operatorname{grade}(M_i) \ge \tau\bigr\}
\end{align}
\noindent\textit{where}  
\(\operatorname{grade}(M_i)\) is the score assigned by the LLM-as-a-Judge grader to answer \(M_i\);  
\(\tau\) is the minimum score required for a “pass”;  
\(N\) and \(\mathbf{1}\{\cdot\}\) are as defined above.

\point{Avg. Latency} The mean time required to generate an answer for a single
test case.

\begin{align}
\mathrm{AvgLatency}
  &= \frac{1}{N} \sum_{i=1}^{N} t_i
\end{align}
\noindent\textit{where} \(t_i\) is the latency for question \(i\) and
\(N\) is the number of questions in the dataset.

\subsubsection{\textbf{Results}}
The experimental results are presented in Tables~\ref{tab:e2e test general}, \ref{tab:e2e test regulation} and \ref{tab:e2e test artifact} across different benchmarks. First, CBA substantially outperforms the vanilla LLM across all quality metrics on all the benchmarks, showing that our compliance-oriented enhancements are effective (e.g., on Compliance Knowledge Benchmark, the best CBA configuration more than doubles the average match rate (83.7\% vs. 41.7\%) and the LLM-judge's pass rate (82.0\% vs. 20.0\%) compared to the vanilla LLM). In fact, the Router mode leads in terms of quality metrics,  proving to be the most performant mode on all benchmarks except the the artifact understanding one (which we find reasonable as this benchmark requires agentic actions for all samples). From the latency perspective, we notice that because the router needs one additional LLM call to determine the appropriate mode to engage for the given user query, it tends to incur longer time compared to directly entering specific modes.  In addition, it is largely expected that \fa on average incurs higher overhead because agent performs multi-step actions which is intrinsically time-consuming, especially for datasets with difficult questions like the Regulation Knowledge Benchmark (Table~\ref{tab:e2e test regulation}). For Compliance Knowledge Benchmark (Table~\ref{tab:e2e test general}), \fa runs surprisingly faster than other setups. After inspecting logs, we found that \fa chose incorrect but fast tools (e.g., keyword-based knowledge retrieval) and generated the answer with only one step, resulting in low latency but also low quality. On the other hand, the questions from this particular dataset (understanding compliance concepts) can be better answered with RAG with a large knowledge base. That's why \ft and router are slightly slower but perform better on this dataset.
Overall, the experimental results suggest that using the router is generally a robust and reliable choice for achieving both high response quality, and acceptable response latency at the same time, across a range of tasks.

\begin{table}[h]
  \centering
  \renewcommand{\arraystretch}{1.5}
  \caption{End-to-end quality and average latency metrics on
           Compliance Knowledge Benchmark (50 samples).}
  \begin{tabular}{|>{\raggedleft\arraybackslash}p{1.7cm}|
                      >{\centering\arraybackslash}p{1.25cm}|
                      >{\centering\arraybackslash}p{1.25cm}|
                      >{\centering\arraybackslash}p{1.25cm}|
                      >{\centering\arraybackslash}p{1.25cm}|}
    \hline
    \multicolumn{1}{|c|}{\textbf{Model}} &
      \textbf{Avg.\ Latency (s)} &
      \textbf{Global Match Rate (\%)} &
      \textbf{Average Match Rate (\%)} &
      \textbf{Pass Rate (\%)}\\ \hline
    Vanilla LLM & 11.38 & 28.1 & 41.7 & 20.0 \\ \hline
    \ft         & 10.50 & 74.3 & 80.3 & 76.0 \\ \hline
    \fa         & \textbf{9.10} & 49.1 & 50.5 & 42.0 \\ \hline
    Router      & 12.24 & \textbf{79.0} & \textbf{83.7} & \textbf{82.0} \\ \hline
  \end{tabular}
  \label{tab:e2e test general}
\end{table}

\begin{table}[h]
  \centering
  \renewcommand{\arraystretch}{1.5}
  \caption{End-to-end quality and average latency metrics on
           Regulation Knowledge Benchmark (14 samples).}
  \begin{tabular}{|>{\raggedleft\arraybackslash}p{1.7cm}|
                      >{\centering\arraybackslash}p{1.25cm}|
                      >{\centering\arraybackslash}p{1.25cm}|
                      >{\centering\arraybackslash}p{1.25cm}|
                      >{\centering\arraybackslash}p{1.25cm}|}
    \hline
    \multicolumn{1}{|c|}{\textbf{Model}} &
      \textbf{Avg.\ Latency (s)} &
      \textbf{Global Match Rate (\%)} &
      \textbf{Average Match Rate (\%)} &
      \textbf{Pass Rate (\%)}\\ \hline
    Vanilla LLM & 12.07 & 32.9 & 32.8 & 42.9 \\ \hline
    \ft         & \textbf{10.43} & 43.4 & 35.7 & \textbf{57.1} \\ \hline
    \fa         & 16.29 & 42.1 & 36.9 & \textbf{57.1} \\ \hline
    Router      & 11.69 & \textbf{50.0} & \textbf{46.1} & \textbf{57.1} \\ \hline
  \end{tabular}
  \label{tab:e2e test regulation}
\end{table}

\begin{table}[h]
  \centering
  \renewcommand{\arraystretch}{1.5}
  \caption{End-to-end quality and average latency metrics on
           Privacy Artifact Understanding Benchmark (54 samples).}
  \begin{tabular}{|>{\raggedleft\arraybackslash}p{1.7cm}|
                      >{\centering\arraybackslash}p{1.25cm}|
                      >{\centering\arraybackslash}p{1.25cm}|
                      >{\centering\arraybackslash}p{1.25cm}|}
    \hline
    \multicolumn{1}{|c|}{\textbf{Model}} &
      \textbf{Avg.\ Latency (s)} &
      \textbf{Global Match Rate (\%)} &
      \textbf{Average Match Rate (\%)}\\ \hline
    Vanilla LLM & \textbf{6.74} & 10.1 & 9.6 \\ \hline
    \ft         & 7.52 & 8.4 & 9.8 \\ \hline
    \fa         & 7.93 & \textbf{60.5} & \textbf{69.7} \\ \hline
    Router      & 9.02 & 58.0 & 67.5 \\ \hline
  \end{tabular}
  \label{tab:e2e test artifact}
\end{table}

\section{Related Work}

While there has been a rich body of work on customizing and tailoring conversational AI systems into enterprise environments~\cite{dong2023towards, muthusamy2023towards}, this is the first work to our knowledge that focuses on designing AI assistants specifically for assisting compliance tasks in the enterprise configuration.
Besides enterprise customization, query routing is another active LLM research area. In recent years, a number of efforts~\cite{hu2024routerbench, hu2024mars} have been made to advance this field, based on various classification mechanisms. This work is again the first to focus on categorizing queries according to their relevance to compliance tasks.

\section{Future Work}

While CBA has shown promising results, there are several areas for future work.
\begin{itemize}
    \item Contextual understanding enhancements: We plan to explore more advanced natural language processing techniques to improve CBA's ability to understand question context.
    \item Knowledge graph expansion: We plan to expand the knowledge graph to include more regulations, guidelines, and best practices to be retrieved.
    \item Agent tool set expansion: We plan to integrate CBA with other tools and systems used by enterprise employees to provide a more seamless experience.
    \item Evaluation improvement: We plan to evaluate CBA in real-world settings to assess its effectiveness in more practical scenarios.
\end{itemize}

\section{Conclusion}

In this paper, we presented Compliance Brain Assistant, a conversational agentic AI assistant designed to help enterprise employees in navigating the complex landscape of compliance. We described the architecture and technical enhancements made to CBA and presented experimental results demonstrating its effectiveness. While there are areas for future work, CBA has the potential to significantly improve the efficiency and accuracy of compliance tasks in enterprise environments.
\begin{acks}
We sincerely thank Mark Harman for his
valuable feedback and thoughtful discussions throughout this work. We appreciate Yuriy Hulovatyy's contributions to CBA's early versions. We also thank Stephen Ou and engineers working with him for supporting CBA to be internally adopted.
\end{acks}

\bibliographystyle{ACM-Reference-Format}
\bibliography{acmart}

\end{document}